\documentclass[manuscript, nonacm]{acmart}
\usepackage[ruled,vlined]{algorithm2e}
\AtBeginDocument{%
  }

\begin{document}

\title{Real-Time Procedural Learning From Experience for AI Agents}

\author{Dasheng Bi}
\email{dbi@altrina.com}
\authornote{Both authors contributed equally to this research.}
\affiliation{%
  \institution{Altrina}
  \city{Menlo Park}
  \state{California}
  \country{USA}
}

\author{Yubin Hu}
\email{harvey@altrina.com}
\authornotemark[1]
\affiliation{%
  \institution{Altrina}
  \city{Menlo Park}
  \state{California}
  \country{USA}
}

\author{Mohammed N. Nasir}
\email{mo@altrina.com}
\affiliation{%
  \institution{Altrina}
  \city{Menlo Park}
  \state{California}
  \country{USA}
}

\renewcommand{\shortauthors}{Bi, Hu, and Nasir}

\begin{abstract}
Learning how to do things from trial and error in real time is a hallmark of biological intelligence, yet most LLM-based agents lack mechanisms to acquire procedural knowledge after deployment. We propose Procedural Recall for Agents with eXperiences Indexed by State (PRAXIS), a lightweight post-training learning mechanism that stores the consequences of actions and retrieves them by jointly matching environmental and internal states of past episodes to the current state. PRAXIS augments agentic action selection with retrieved state-action-result exemplars that are generated in real time. When evaluated on the REAL web browsing benchmark, PRAXIS improves task completion accuracy, reliability, and cost efficiency across different foundation model backbones, and shows preliminary generalization to unseen tasks in similar environments. These results demonstrate that PRAXIS enables the practical adoption of AI agents in fast-evolving stateful environments by helping them learn new procedures effectively.
\end{abstract}




\maketitle

\section{Introduction}
\subsection{AI Agents and Procedural Learning}

AI agents are artificial intelligence systems capable of observing and taking actions in an environment. As adoption spreads across industries, there is a growing need for AI agents to quickly learn domain- or user-specific information. There are two main classes of information that agents are typically expected to learn:
\begin{enumerate}
    \item \textbf{Facts}: atomic pieces of information independent of the state of the agent or environment (\emph{e.g.}, a user's name, preferences, organizational charts). Facts can change over time, but at any given moment they are generally context‑independent.
    \item \textbf{Procedures}: established conventions for doing things (\emph{e.g.}, "how to troubleshoot a failed login" or "how to guide a customer through a sales process to the most fitting product"). Procedures can be viewed as a sequence of state-dependent requirements or preferences over actions. 
\end{enumerate}

In real-world applications, learning and optimizing procedures in real time are at least as important as learning facts. While frameworks such as Mem0 \citep{chhikara_mem0_2025} and Letta \citep{packer2024memgptllmsoperatingsystems} focus on long-term \emph{factual} memory, effective post-training learning of \emph{procedures} in AI agents remains relatively underexplored. 

A na\"ive approach is \emph{a priori} procedural specification: a human writes rules or standard operating procedures (SOPs) that are included in the agent's context at inference time. This approach effectively reduces procedures to a large bundle of facts. In practice, this approach faces challenges because (1) many procedures are not fully documented, as humans are often trained by observation rather than by reading long SOPs; (2) enumerating all states and edge cases in a combinatorial space is difficult; and (3) procedures can become obsolete quickly as environments change. We argue that a more robust approach is to learn procedures \emph{a posteriori} from demonstrations or experience. Inspired by state-dependent memory in psychology \citep{tulving_encoding_1973, bower_mood_1981}, we propose PRAXIS, a concrete method for procedural recall and show that it improves agent accuracy, reliability, and efficiency in web browsing settings. Our method is compatible with both experiences demonstrated by a human expert and actual trajectories generated by the AI agent itself.

\subsection{Web Agents and the Browser Environment}

Human-facing web applications almost always require multi-step interactions to accomplish meaningful goals (\emph{e.g.}, purchasing an item online requires searching, filtering, logging in, completing forms, and checking out)---a procedure. Moreover, these procedures must also adapt to changing environments (\emph{e.g.}, an e-commerce site may have seasonal pop-ups or redesigned interfaces), making web browsing a natural environment to study \emph{a posteriori} procedural learning. Importantly, even when tasks are obvious to humans, comprehensive procedures are rarely documented, and high personalization limits pretraining coverage in foundation models. As AI-based design tools increasingly generate and update web platforms, the economic value shifts to novel, previously unseen interfaces and interaction flows, pushing agents into out-of-distribution states and rendering \emph{a priori} SOPs brittle. A post-training, state-indexed procedural memory thus becomes essential for robust web automation, allowing agents to acquire and reuse procedures precisely when new states appear.

\section{Related Work}
\paragraph{External Memory for AI Chatbots}
A broad class of systems augment LLMs with non-parametric memory in the conversational environment. Retrieval-augmented generation (RAG) attaches a document store to provide factual knowledge at inference time \citep{lewis_retrieval-augmented_2021}. In agentic settings, persistent memory frameworks such as Letta (formerly MemGPT) \citep{packer2024memgptllmsoperatingsystems} provide hierarchical storage and dynamic context management. Mem0 \citep{chhikara_mem0_2025} provides a queryable, cross-session memory for user preferences and long-range conversational context. Academic frameworks include MemoryBank \citep{zhong_memorybank_2023} which mimics human long-term memory with continual decay and reinforcement, and A-MEM \citep{xu_-mem_2025} which dynamically links and evolves structured notes. These approaches typically focus on factual memory for agents in conversational environments. In contrast, our method focuses on learning action policies in stateful visual environments that are significantly more complex and not entirely observable like the web environment.

\paragraph{Experience‑Based Self‑Improvement and Workflow Memories}
A complementary line of work improves agents via self‑reflection. Reflexion \citep{shinn_reflexion_2023} maintains verbal reflections in an episodic buffer to guide subsequent trials; Self‑Refine \citep{madaan_self-refine_2023} iteratively critiques and edits its own outputs; and CLIN \citep{majumder_clin_2023} performs continual task adaptation with a persistent textual memory of causal abstractions. These methods are effective but are not well-tested in visual environments and generally do not encode information of environmental state. There has also been a line of work on experience- or workflow-based memories for agents. Agent Workflow Memory \citep{wang_agent_2024}, Synapse \citep{zheng_synapse_2024}, and ExpeL \citep{zhao_expel_2023} induce abstracted, natural language workflows from successful trajectories and retrieve them to augment prompts at test time. In contrast, our method performs local state-based recall that is grounded primarily in the live environment state, a factor not present in prior works, and secondarily to the goal. Moreover, we index memories with explicit state and action descriptors, rather than high-level trajectories, enabling precise recall and learning of minute details required for environments like the web.

\section{Methods}

\subsection{Altrina Agent}

The experiments in this study were conducted with Altrina\footnote{The agent system described in this work was previously known as Tessa in prior publications \citep{TessaAITeam2025a, TessaAITeam2025b}.}, a frontier AI agent operating in computer use environments. Altrina is capable of perceiving the environment both visually and as compressed textual information, taking actions like a human on a computer, and executing complex directives end-to-end. While Altrina's capabilities extend beyond acting in the browser, we restrict the action space to web-related actions for this study. In the browser environment, we previously showed that the baseline Altrina agent achieves state-of-the-art results on multiple benchmarks \citep{TessaAITeam2025a, TessaAITeam2025b}, including WebVoyager \citep{he_webvoyager_2024} and REAL \citep{garg_real_2025}.

Altrina's scaffolding layer orchestrates underlying foundational models with a "node-based" architecture. Each node is designed for a specific function in the agentic loop. In this study, we modify the \emph{action selection node}, which is responsible for deciding what the next agent action should be given the current state and progress towards the task objective, by adding a dedicated \emph{procedural memory} section to its context.

\subsection{State-Dependent Memory}

Mirroring state-dependent memory in psychology, which finds improved recall when internal state and external context at retrieval match those present during encoding \citep{tulving_encoding_1973, bower_mood_1981}, we designed a state-dependent memory for Altrina in which indexing and retrieval are based on the browser's environment state and the agent's internal state. 

Each memory entry contains the following components:
\begin{enumerate}
    \item $M_i^{\text{env-pre}}$, a description of the environmental state in which this memory is generated
    \item $M_i^{\text{int}}$, the internal state of the agent at the time of the experience, including the overall directive the agent is trying to achieve
    \item $a_i$, the action taken at the time
    \item $M_i^{\text{env-post}}$, the state of the environment after the action was taken
\end{enumerate}

Formally, we use the following procedure (Alg. \ref{alg:pm_retrieve}) to retrieve a set of memories that may be informative to the agent:

\begin{algorithm}[h]
\caption{Procedural Memory Retrieval}
\label{alg:pm_retrieve}
\KwIn{Memory environment states $\{M^{\text{env}}_i\}_{i=1}^n$, query environment state $Q^{\text{env}}$, memory internal states $\{M^{\text{int}}_i\}_{i=1}^n$, query internal state $Q^{\text{int}}$, internal state embedding function $f$, search breadth $k$, similarity threshold $\tau$}
\KwOut{Retrieved memory indices $\mathcal{R}$}

\For{$i \gets 1$ \KwTo $n$}{
  $v_i \leftarrow \textsc{IoU}(M^{\text{env}}_i, Q^{\text{env}})$\;
  $\ell_m \leftarrow |M^{\text{env}}_i|,\ \ell_q \leftarrow |Q^{\text{env}}|$\;
  $l_i \leftarrow \textsc{LengthOverlap}(\ell_m, \ell_q) =
    1 - \dfrac{|\ell_m - \ell_q|}{\max(\ell_m, \ell_q)}$\;
  $s^{\text{env}}_i \leftarrow v_i \cdot l_i$\;

  $s^{\text{int}}_i \leftarrow \left\langle f(M_i^{\text{int}}), f(Q^{\text{int}})\right\rangle$;\
}
$\mathcal{R}^{\text{env}} \leftarrow \textsc{TopkIndices}(s^{\text{env}}, k)$\;
$\widetilde{\mathcal{R}} \leftarrow \textsc{Sorted}(\mathcal{R}^{\text{env}};\ \text{key}=i\mapsto s^{\text{int}}_i;\ \text{descending})$\;
$\mathcal{R} \leftarrow [\, i \in \widetilde{\mathcal{R}} \mid s^{\text{env}}_i \ge \tau \,]$\;
\Return $\mathcal{R}$\;
\end{algorithm}

\subsection{Benchmark}

A number of benchmarks exist to evaluate the performance of AI agents. MiniWoB/MiniWoB++ target short-horizon GUI primitives \citep{liu_reinforcement_2018}; Mind2Web and WebVoyager target relatively straightforward workflows on live websites \citep{deng_mind2web_2023, he_webvoyager_2024}; WebArena hosts replicas of a few websites for local testing of relatively simple tasks with text-based evaluations \citep{zhou_webarena_2024}. To ensure reproducible evaluation on tasks of real-world relevance, we evaluate on deterministic clones of functional websites using the REAL benchmark \citep{garg_real_2025}. REAL provides replicas of 11 commonly used sites and 112 everyday tasks of varying complexity, along with both programmatic state checks for action tasks and rubric-guided natural language evaluations for information retrieval tasks. Prior benchmarking reports that frontier models with na\"ive scaffolding achieve at most $\sim\!41\%$ success, leaving substantial room for post-training procedural learning.

\section{Results}

\subsection{Procedural memory improves agent accuracy}

\begin{figure}[h]
    \centering
    \includegraphics[width=0.75\linewidth]{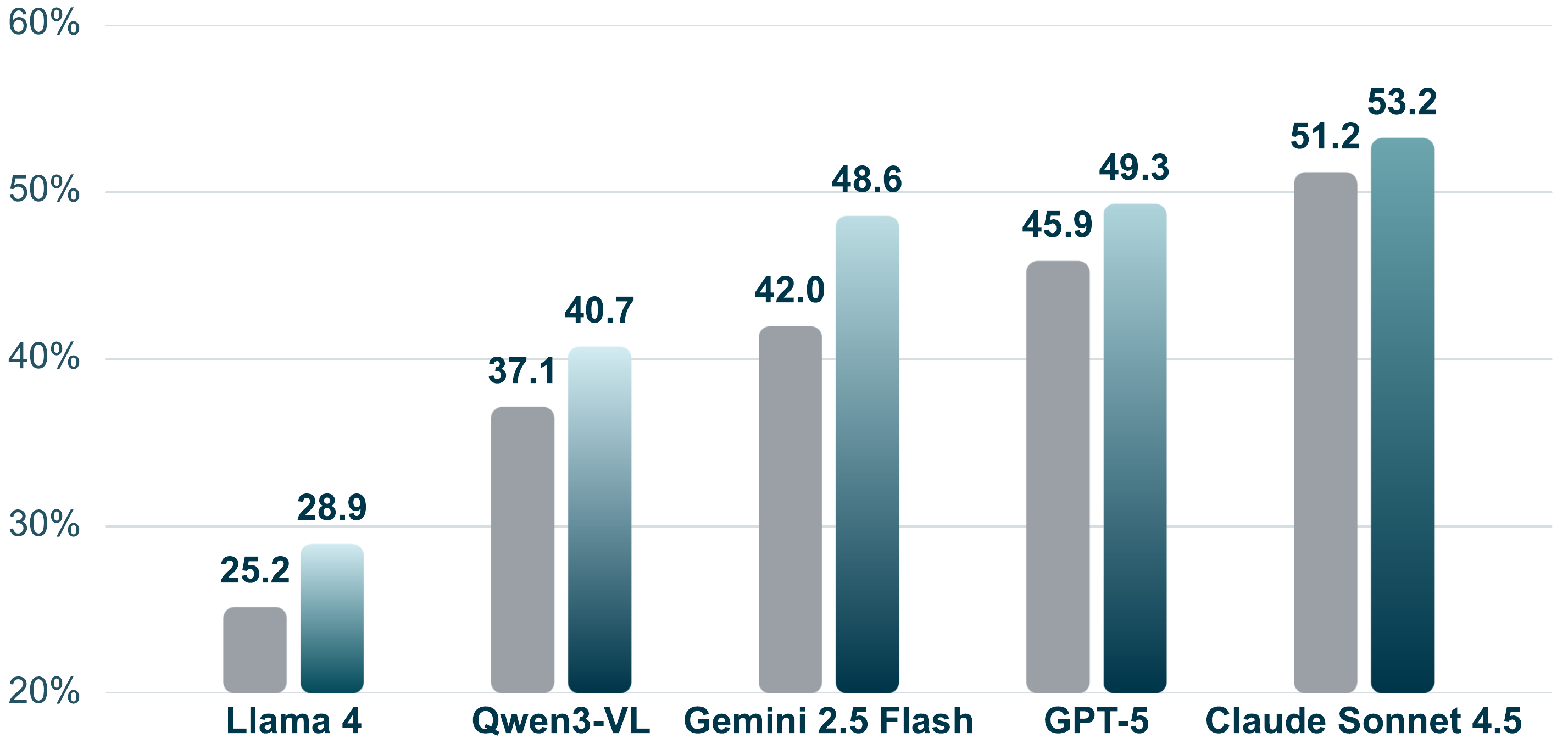}
    \caption{Performance on REAL benchmark for different models with (blue) and without (grey) procedural memory.}
    \label{fig:fig1}
    \Description{A bar graph showing overall agent performance on REAL benchmark for different models with and without procedural memory.}
\end{figure}

We benchmarked Altrina on REAL using several VLM backbones for each of the agentic compute nodes, either with or without access to procedural memory (Fig.~\ref{fig:fig1}). Procedural memory consistently improved the performance of Altrina (average performance across models increased from $40.3\%$ to $44.1\%$, mean accuracy over five repetitions).

We also measured the best-of-5 accuracy (Table~\ref{tab:tab1}), which reflects the agent’s capability frontier under repeated attempts. Procedural memory also improved best-of-5 accuracy (average across models from $53.7\%$ to $55.7\%$). Together, these results suggest that retrieved procedural memory traces provide reusable priors that improve agentic behavior on complex tasks, and that such memory traces may be generalized across similar tasks, supporting the completion of novel tasks.

\begin{table}[h]
\caption{Best-of-5 accuracy over tasks.}
\label{tab:tab1}
\begin{tabular}{lcc}
\toprule
Model & No PM & With PM \\
\midrule
Llama 4 & 47.3 & \textbf{52.7} \\
Qwen3-VL & 44.6 & \textbf{47.3} \\
Gemini 2.5 Flash & 59.8 & \textbf{61.6} \\
GPT-5 & 56.2 & \textbf{57.1} \\
Claude Sonnet 4.5 & \textbf{60.7} & {59.8} \\
\bottomrule
\end{tabular}
\end{table}

\subsection{Procedural memory improves agent reliability}

\begin{figure}[h]
    \centering
    \includegraphics[width=0.75\linewidth]{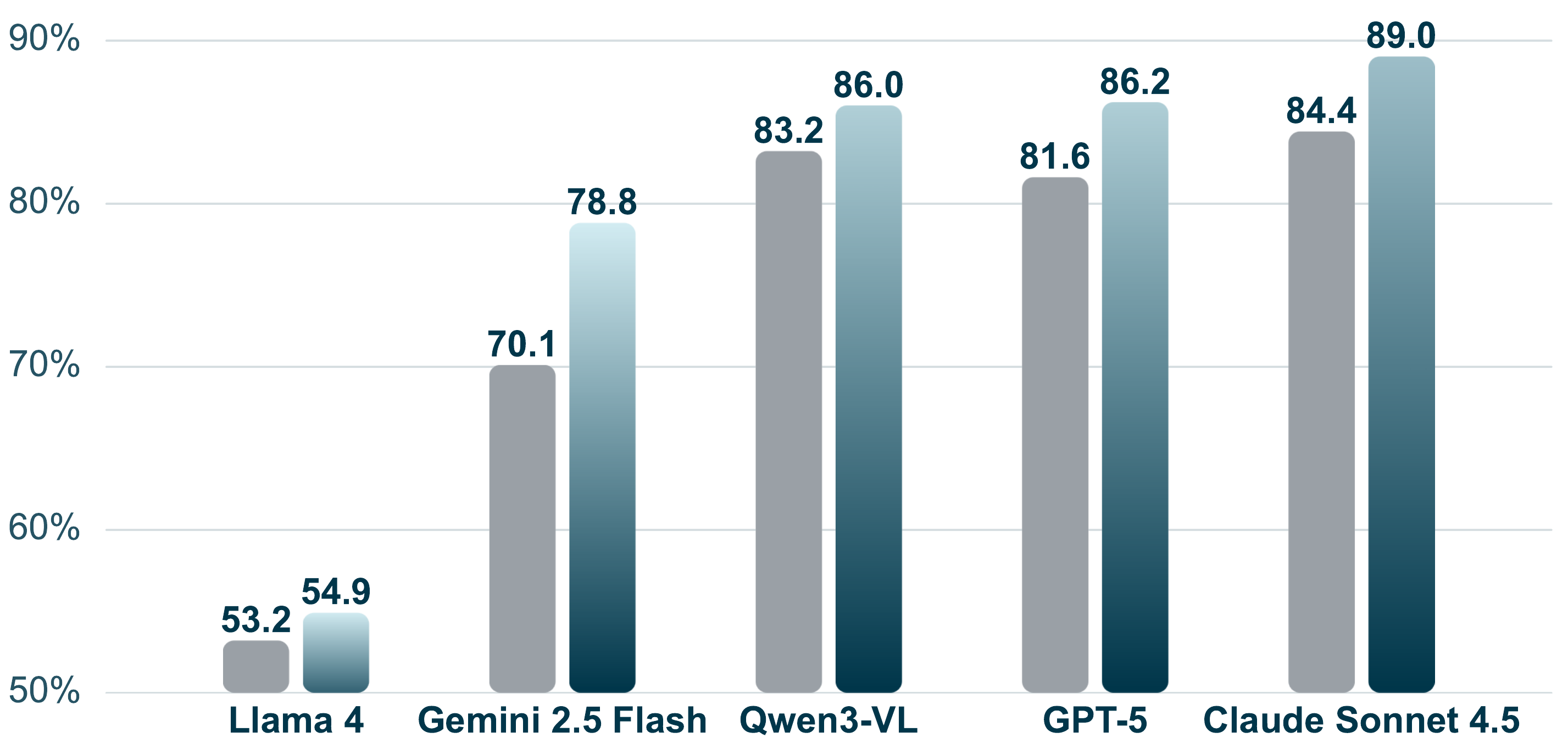}
    \caption{Agent reliability on REAL benchmark for different models with (blue) and without (grey) procedural memory.}
    \label{fig:fig2}
    \Description{A bar graph showing agent reliability on REAL benchmark for different models with and without procedural memory.}
\end{figure}

We define \emph{reliability} as the mean success rate over five repetitions of a task, averaged over all tasks in REAL having at least one successful run over the five repetitions. Procedural memory 
improved the reliability of Altrina from $74.5\%$ to $79.0\%$ averaged across all models (Fig.~\ref{fig:fig2}). These results suggest that retrieved procedural memory traces can suppress unwanted stochastic variance in the underlying vision-language models by biasing their decisions towards previously successful trajectories under similar states, thereby improving the reliability and repeatability of the high-level agentic behavior.

\subsection{Procedural memory improves agent efficiency}

We measure \emph{efficiency} as the average number of steps taken on tasks in REAL with at least one successful run across five repetitions. Procedural memory reduces steps-to-completion on average from $25.2$ to $20.2$ across models (Table~\ref{tab:tab2}). This indicates that procedural memory recall effectively steers agentic behavior along both correct and more direct trajectories.

\begin{table}[h]
\caption{Average number of steps to complete task.}
\label{tab:tab2}
\begin{tabular}{lcc}
\toprule
Model & Base & With PM \\
\midrule
Llama 4 & 19.8 & \textbf{16.2} \\
Qwen3-VL & 27.7 & \textbf{20.8} \\
Gemini 2.5 Flash & 28.9 & \textbf{22.3} \\
GPT-5 & {24.2} & \textbf{20.7} \\
Claude Sonnet 4.5 & {25.2} & \textbf{21.0} \\
\bottomrule
\end{tabular}
\end{table}

\subsection{Procedural memory performance scales with retrieval breadth}

\begin{figure}[h]
    \centering
    \includegraphics[width=0.75\linewidth]{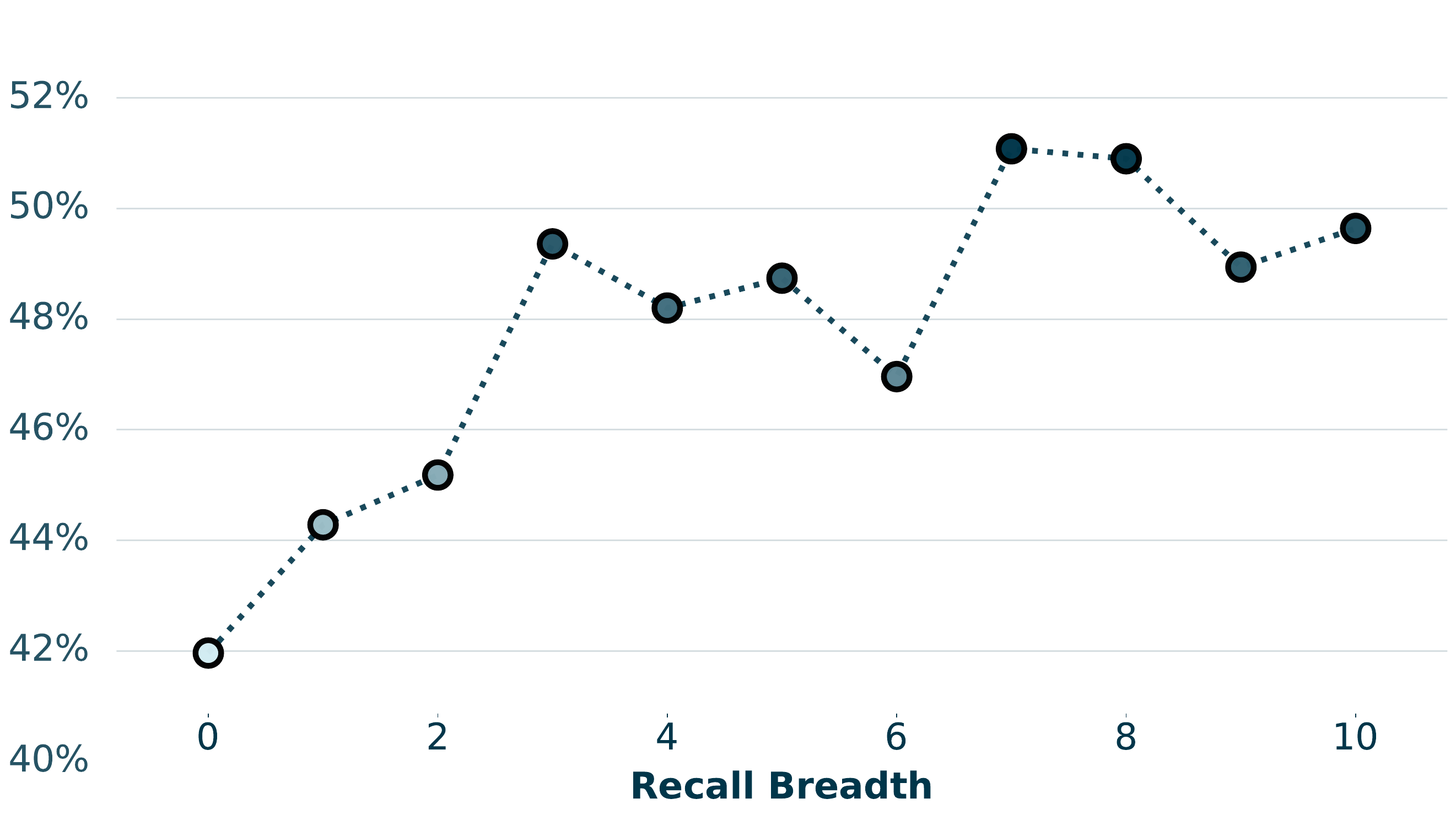}
    \caption{Performance as a function of retrieval breadth.}
    \label{fig:fig3}
    \Description{A line graph showing agent performance on REAL benchmark with different recall breadths for procedural memory.}
\end{figure}

We ablated the retrieval breadth hyperparameter $k$ in Alg.~\ref{alg:pm_retrieve} to study its effect on the performance of procedural memory (Fig.~\ref{fig:fig3}). Due to resource limitations, we only tested the effects of this ablation on the agent with a Gemini 2.5 Flash backbone. We found that performance generally increased in steps as we broadened our retrieval breadth, with slight decreases within each step, and converged to a plateau, suggesting that while there exists some local context crowding, procedural memory offers increasingly helpful and generalizable context to the language model at a larger scale.

\section{Discussion}
\paragraph{Personalized learning as a critical component of AI agents in the economy}
We envision a future where, instead of replacing humans, AI agents work alongside them. To encompass the diverse activities across our economy and maintain high collaboration efficiency, users will need to customize their AI agent with their own data and procedures. Even if it becomes possible to quickly and efficiently train everything into a single, universal model, we may not wish to, as each user should be able to decide whether to share their private knowledge with the world. In this context, personalized learning methods like PRAXIS that customize agents, not only on a superficial level, but also in terms of real capabilities, will be critical to the adoption of agents in the economy. 

\paragraph{Summary of contributions}
This paper introduces \emph{state-dependent memory}, an \emph{a posteriori} learning mechanism that stores local interaction traces and retrieves them by jointly matching the \emph{environment state} and the agent’s \emph{internal objective}. When tested as an integration into the Altrina web agent, our method yields consistent improvements on the REAL web browsing benchmark across diverse VLM backbones: higher average accuracy, higher best‑of‑5 accuracy, better reliability, and fewer steps to completion. An ablation shows increased gains with retrieval breadth $k$. Together, our results indicate that state-dependent memory provides reusable local state-to-action priors that guide AI agents towards robust, generalizable behavior.

\paragraph{Future directions}
\emph{Beyond web environments.} State-dependent memory is conceptually agnostic to the environment, and the same idea can be naturally extended to general cases of agentic computer use.
\emph{Richer state encoding.} Our proof-of-concept implementation of state-dependent memory uses basic visual and DOM feature overlap along with simple similarity metrics. A richer encoder can improve both retrieval quality and invariance to superficial changes.
\emph{Adaptive retrieval mechanisms.} Rather than a fixed retrieval based on state similarity heuristics, the retrieval mechanism can account for real-time factors such as uncertainty and compute budget. For highly uncertain scenarios, retrieval can also be iterative to improve quality.
\emph{From action agents to alignment agents} While the training signal in this paper is objective task success or failure, PRAXIS can use user preference as a training signal in cases where no objective standard exists. This would be accomplished by observing the user's inputs and feedback over time, steering the actions accordingly at each iteration, and converging to a procedural memory that encodes the user's preferences for how a task is done.

\bibliographystyle{ACM-Reference-Format}
\bibliography{references}

\end{document}